\documentclass[conference]{IEEEtran}
\usepackage{cite}
\usepackage{amsmath,amssymb,amsfonts}
\usepackage{graphicx}
\usepackage{xcolor}

\newif\ifshowrevisions
\showrevisionstrue

\usepackage{booktabs}
\usepackage{tikz}
\usetikzlibrary{arrows.meta,positioning,shapes.geometric,calc}
\usepackage{placeins}
\usepackage{url}
\newcommand{\clipfn}{\operatorname{clip}}

\usepackage{fancyhdr}

\fancypagestyle{firstpage}{
    \fancyhead[L]{\footnotesize ©
    This paper is accepted at the 2027 International Conference
    on Artificial Intelligence and Applied Intelligent Systems
    (AIAIS 2027), January 22--24, 2027, Hong Kong.}
    \fancyhead[R]{}
}

\begin{document}
\thispagestyle{firstpage}

\title{Client-Aware Security Adaptive Early Exit Vision Transformers for Heterogeneous Federated Learning}

\author{
\IEEEauthorblockN{
Sahar Refaei Ghalibafan\textsuperscript{1},
Samira Nazari\textsuperscript{2},
Christian Herglotz\textsuperscript{1},
Maksim Jenihhin\textsuperscript{3},
Mahdi Taheri\textsuperscript{3,4}
}
\IEEEauthorblockA{
\textsuperscript{1}Brandenburg University of Technology, Cottbus, Germany\\
\textsuperscript{2}University of Zanjan, Zanjan, Iran\\
\textsuperscript{3}Tallinn University of Technology, Tallinn, Estonia\\
\textsuperscript{4}Humboldt University of Berlin, Berlin, Germany\\ mahdi.taheri@hu-berlin.de}}

\maketitle
\thispagestyle{firstpage}

\begin{abstract}
Federated learning (FL) trains a shared model across data-owning clients without centralizing raw data, but client data difficulty and compute budgets can differ. We study local early-exit Vision Transformer (ViT) inference, where an intermediate classifier may return a prediction before the final Transformer stage. The proposed method replaces one global acceptance threshold with a client-specific base threshold from label entropy and shallow-exit confidence, then adjusts each input using predictive entropy, confidence, top-two class margin, and exit depth. Aggregation uses only clients that updated a parameter. 
Security-aware routing reaches 88.70\% CIFAR-10 accuracy with 32.0\% depth saving and 73.7\% ImageNet-1K top-1 accuracy with 30\% saving. On matched six-client non-IID CIFAR-100, full-depth accuracy is 72.28\% versus 68.20--71.80\% for four heterogeneous-FL baselines; robustness-aware routing reaches 70.01\% with 32.55\% saving. 
\end{abstract}

\section{Introduction}
Federated learning (FL) trains a shared model by exchanging model updates between a central server and distributed \emph{clients}, where each client owns local data that are not uploaded \cite{mcmahan2017, taheri2025rl, hasa}. Clients can differ in hardware, sample count, label distribution, and task difficulty. In this work, training is server-coordinated, but inference is performed locally on the client/edge device. The first gap is that heterogeneous training does not itself define a client-aware inference rule.

A Vision Transformer (ViT) processes an image through sequential Transformer blocks \cite{dart, fusion}. An \emph{early-exit network} adds intermediate classifiers so an input may stop before the final block \cite{teerapittayanon2016}. An \emph{acceptance threshold} specifies the confidence required to stop; otherwise the input is \emph{deferred}, meaning it continues to the next, deeper exit. Most systems use one fixed threshold or one globally shared threshold schedule. The second gap is that such policies adapt to the input but not to the client from which it originates.

Confidence alone is incomplete because neural networks can be miscalibrated \cite{guo2017}, and adversarially perturbed inputs can remain confidently wrong \cite{goodfellow2015,madry2018}. Selective prediction therefore ranks predictions by uncertainty so difficult inputs can be deferred \cite{geifman2017,corbiere2019}. Confidence, predictive entropy, and top-two class margin capture different aspects of that uncertainty. The third gap is a clear connection between these sample signals, client difficulty, and when deferral to a deeper exit actually reduces error.

We address the three gaps with (i) a client-difficulty score from label entropy and shallow-exit confidence, (ii) sample-adaptive routing from confidence, entropy, class margin, and exit depth with a conditional-error analysis, and (iii) coverage-aware aggregation that averages a parameter only across clients that trained it. Section~II reviews related work, Section~III defines the method and notation, Section~IV evaluates it, and Section~V concludes.

\section{Related Work}
HeteroFL changes neural-network \emph{width}, i.e., active channels \cite{diao2021}; DepthFL changes model \emph{depth}, i.e., trainable sequential blocks \cite{kim2023}; and ScaleFL combines width/depth scaling, exits, and self-distillation \cite{ilhan2023scalefl}. ReeFL shares a recurrent exit classifier across depths and selects a client-specific distillation teacher \cite{lee2024}. FedEE uses personalized early-exit blocks for uncertainty-aware healthcare FL \cite{zhang2025fedee}. Here, \emph{personalization} means that client statistics change the inference threshold rather than only the trained model capacity.

Early exiting stops at an intermediate classifier when an acceptance condition is met \cite{teerapittayanon2016}. Token reduction instead removes redundant ViT tokens; Byun \emph{et al.} combine both mechanisms using dataset profiling and intermediate confidence \cite{byun2026}. Maximum softmax confidence can be miscalibrated \cite{guo2017}; selective classification therefore studies prediction versus deferral using risk--coverage behavior \cite{geifman2017}, with entropy and class margin providing complementary uncertainty signals \cite{corbiere2019}.

An \emph{adversarial perturbation} is a small, intentionally chosen input change that increases model error. Fast Gradient Sign Method (FGSM) and Projected Gradient Descent (PGD) are gradient-based attacks \cite{goodfellow2015,madry2018}; broader suites can include Carlini--Wagner and AutoAttack \cite{carlini2017,croce2020}. 

The remaining intersection is a federated multi-exit system that adapts inference at both the client and sample levels while correctly aggregating unequal model-prefix coverage.

\section{Methodology}
\subsection{Problem Formulation and Terminology}
Consider $K$ federated clients indexed by $k\in\{1,\ldots,K\}$. Client $k$ owns the local dataset $\mathcal{D}_k=\{(x_i,y_i)\}_{i=1}^{n_k}$, where $n_k$ is its number of samples, $x_i$ is an input image, and $y_i\in\{1,\ldots,C\}$ is its label among $C$ classes. The server maintains the global model parameter vector $\boldsymbol{\theta}^{t}$ after communication round $t$, where one \emph{communication round} consists of a server broadcast, local client training, and server aggregation; $t\in\{0,\ldots,T_{\mathrm{FL}}-1\}$ and $T_{\mathrm{FL}}$ is the total number of FL rounds.

The shared ViT contains $L$ ordered Transformer stages and one classifier after each stage. The exit index is $e\in\{1,\ldots,L\}$. For input $x$, stage function $f_e$ maps the previous representation $h_{e-1}(x)$ to $h_e(x)$, and exit-head parameters $(W_e,b_e)$ produce the class-probability vector $p_e(x)\in[0,1]^C$:
\begin{equation}
 h_e(x)=f_e(h_{e-1}(x)),\qquad
 p_e(x)=\operatorname{softmax}(W_eh_e(x)+b_e).
\end{equation}
The softmax function converts the exit-head scores into class probabilities. For class index $c$, $p_{e,c}(x)$ is its predicted probability. The exit prediction is $\hat y_e(x)=\arg\max_c p_{e,c}(x)$, and $C_e(x)=\max_c p_{e,c}(x)$ is the maximum prediction confidence. \emph{Routing} is the sequential inference procedure that either accepts the current exit or defers the input to exit $e+1$. A client \emph{subnetwork} (subnet) is the shared-model prefix ending at the deepest stage that the client is allowed to train.

\subsection{Local Multi-Exit Training}
Let $m_k\in\{1,\ldots,L\}$ be the deepest trainable exit assigned to client $k$. The client updates the shared prefix through stage $m_k$ and every exit head up to that depth. Let $\ell_e$ denote the supervised classification loss at exit $e$; it is cross entropy for ordinary labels and the corresponding soft-label loss under MixUp, an augmentation that interpolates training examples and labels. We define the weighted supervised loss
\begin{align}
 \mathcal{L}_{\mathrm{sup}}&=\sum_{e=1}^{m_k}\rho_e\ell_e,\notag\\
 \rho_e&=\frac{r_e}{\sum_{j=1}^{m_k}r_j}.
\end{align}
where $j$ is a summation index and $r_e$ is the unnormalized exit weight. We use linearly increasing base weights $r_e=e$ and double the deepest active weight, i.e., $r_{m_k}=2m_k$, so that shallow heads are trained without allowing their larger count to dominate the local objective.

Knowledge distillation (KD) transfers information from a deeper classifier to a shallower one. The deepest active exit is used as a detached teacher, giving
\begin{equation}
 \mathcal{L}_{\mathrm{KD}}=\frac{T_{\mathrm{KD}}^2}{m_k-1}\sum_{e=1}^{m_k-1}
 D_{\mathrm{KL}}\!\left(p_{m_k}^{(T_{\mathrm{KD}})}\,\|\,p_e^{(T_{\mathrm{KD}})}\right),
\end{equation}
where $D_{\mathrm{KL}}$ is Kullback--Leibler divergence, $T_{\mathrm{KD}}$ is the distillation temperature, and $p_e^{(T_{\mathrm{KD}})}$ is the softmax distribution computed with that temperature. The complete local training loss is $\mathcal{L}_k=\mathcal{L}_{\mathrm{sup}}+\gamma\mathcal{L}_{\mathrm{KD}}$, where $\gamma$ weights the KD term. Thus, $\mathcal{L}_k$ is the loss optimized during local client training; it is not used during inference.

\subsection{Client-Level Difficulty and Base Threshold}
Client-level adaptation uses both a data statistic and a model-behavior statistic because either one alone can miss a source of difficulty. Let $q_{k,c}$ be the fraction of client-$k$ samples with class $c$. The normalized label entropy
\begin{equation}
 H_k=-\frac{1}{\log C}\sum_{c=1}^{C}q_{k,c}\log(q_{k,c}+\varepsilon)
\end{equation}
measures label diversity, where $\varepsilon>0$ is a small numerical constant that prevents $\log 0$. High $H_k$ means the client covers many classes, but it does not indicate whether the current model already predicts that client confidently.

We therefore also measure shallow-exit confidence. Let $P$ be the number of shallow exits used for profiling; the reported experiments use $P=2$. The mean shallow confidence is
\begin{equation}
 \bar C_k=\frac{1}{Pn_k}\sum_{x\in\mathcal{D}_k}\sum_{e=1}^{P}C_e(x),
\end{equation}
and $U_k=1-\bar C_k$ is the client's shallow-confidence deficit. We min--max normalize $H_k$ and $U_k$ across the profiled clients to $[0,1]$, yielding $\widetilde H_k$ and $\widetilde U_k$. Their weighted combination
\begin{equation}
 S_k=0.65\widetilde H_k+0.35\widetilde U_k
\end{equation}
is the \emph{client-difficulty score}. The weights are fixed design choices: label-distribution heterogeneity receives the larger weight, while model confidence supplies a complementary difficulty signal; they are not learned or tuned on the test set.

For scalar $z$ and lower/upper bounds $l,u$, define $\clipfn(z,l,u)=\min(u,\max(l,z))$. The client-specific \emph{base acceptance threshold} is
\begin{equation}
 \tau_k=\clipfn(0.6+0.25S_k,0.45,0.95).
\end{equation}
Because $S_k\in[0,1]$, the coefficient 0.25 limits the unclipped client-specific threshold shift to at most 0.25. A higher $S_k$ therefore makes early acceptance stricter. Clients are profiled once before the first communication round.

\subsection{Sample-Level Uncertainty and Adaptive Exit Rule}
Client difficulty cannot distinguish easy and difficult inputs from the same client. At exit $e$, we therefore compute three sample-level uncertainty signals. First, normalized predictive entropy
\begin{equation}
 H_e(x)=-\frac{1}{\log C}\sum_{c=1}^{C}p_{e,c}(x)\log(p_{e,c}(x)+\varepsilon)
\end{equation}
measures how spread out the predicted class probabilities are. Second, $1-C_e(x)$ is the confidence deficit. Third, let $p_{e,(1)}(x)$ and $p_{e,(2)}(x)$ denote the largest and second-largest class probabilities; their margin
\begin{equation}
 M_e(x)=p_{e,(1)}(x)-p_{e,(2)}(x)
\end{equation}
measures separation between the two leading classes. These signals are complementary: entropy uses the full distribution, confidence uses its highest probability, and margin measures ambiguity between the two most likely classes.

The global resource-aware policy uses
\begin{equation}
 Q_e(x)=H_e(x)[1-C_e(x)],
\end{equation}
a multiplicative sample-difficulty term that becomes large only when both predictive entropy and confidence deficit are high. The robustness-aware policy additionally uses
\begin{equation}
 R_e(x)=0.4H_e(x)+0.4[1-C_e(x)]+0.2[1-M_e(x)],
\end{equation}
where entropy and confidence receive equal primary weight and the class-margin deficit supplies complementary information. These coefficients are fixed implementation choices rather than learned parameters.

All exit policies are instances of
\begin{align}
\tau_{k,e}(x)=\clipfn\!\bigg(&\tau_0+aS_k+bQ_e(x)+cR_e(x)\notag\\
&-d\frac{e}{L},\,\tau_{\min},\tau_{\max}\bigg),
\label{eq:threshold-family}
\end{align}
where $\tau_0=0.6$ is the global baseline threshold; $a,b,c\ge0$ weight client difficulty, multiplicative sample difficulty, and multi-signal sample uncertainty; $d\ge0$ controls \emph{depth relaxation}; $e/L$ is relative exit depth; and $\tau_{\min}=0.45$, $\tau_{\max}=0.95$ are clipping bounds. Depth relaxation means that, after more computation has already been spent, the threshold is gradually reduced so a later exit can accept an input more easily. The first exit satisfying $C_e(x)>\tau_{k,e}(x)$ returns $\hat y_e(x)$; if none accepts the input, exit $L$ is the deterministic final fallback.

\begin{table}[!t]
\caption{Instantiated Exit Policies}
\label{tab:policies}
\centering\scriptsize
\begin{tabular}{lcccc}\toprule
Policy & $a$ & $b$ & $c$ & $d$\\\midrule
Fixed & 0 & 0 & 0 & 0\\
Global resource-aware & 0 & .25 & 0 & .10\\
Client-specific & .25 & 0 & 0 & 0\\
Robustness-aware & 0 & .25 & .20 & .10\\
Combined client+robust. & .25 & .25 & .20 & .10\\\bottomrule
\end{tabular}
\end{table}
Table~\ref{tab:policies} includes the fully combined parameterization to make the policy family explicit. The correction magnitudes 0.25, 0.20, and 0.10 are fixed design values chosen so client/sample corrections remain comparable in scale while depth relaxation is smaller; they are not learned from test data. The reported common global-test ablation isolates fixed, global resource-aware, and robustness-aware routing; client-specific variants require client identity and are evaluated on client-local data rather than assigned one pooled global-test result.

\subsection{Why Deferral Can Reduce Error}
A stricter threshold is not itself an adversarial defense: a perturbed input can remain confidently wrong. The routing rule is useful only when it identifies inputs for which a deeper exit has lower error. Let $A_e$ denote the subset accepted at exit $e$ and $D_e$ the subset deferred from exit $e$ to $e+1$. Let $\Pr(B)$ be the probability (or empirical fraction) of subset $B$, and let $\mathcal{E}_{j\mid B}$ be the conditional misclassification rate of exit $j$ on subset $B$. A two-stage routing policy has risk
\begin{equation}
 \mathcal{R}=\Pr(A_e)\mathcal{E}_{e\mid A_e}+\Pr(D_e)\mathcal{E}_{e+1\mid D_e}.
\end{equation}
If the same deferred samples were forced to exit at $e$, the risk difference would be
\begin{equation}
 \mathcal{R}_{\mathrm{shallow}}-\mathcal{R}
 =\Pr(D_e)\left(\mathcal{E}_{e\mid D_e}-\mathcal{E}_{e+1\mid D_e}\right).
\end{equation}
Deferral reduces expected error exactly when $\mathcal{E}_{e+1\mid D_e}<\mathcal{E}_{e\mid D_e}$. Confidence, predictive entropy, and class margin are therefore routing heuristics for identifying potentially recoverable shallow errors, not correctness certificates.

\subsection{Coverage-Aware Federated Aggregation}
The global model vector $\boldsymbol{\theta}^{t}$ contains individual trainable parameters $\theta_i^t$, where $i$ indexes a scalar or tensor parameter. Let $\mathcal{S}_t$ be the set of clients selected in round $t$, and let $\mathcal{K}_i^t\subseteq\mathcal{S}_t$ be the subset whose active subnet contains and updates parameter $\theta_i$. After local training, $\theta_{i,k}^{t+1}$ denotes client $k$'s updated value for that parameter. We aggregate only clients that covered it:
\begin{equation}
\label{eq:coverage}
\theta_i^{t+1}=\begin{cases}
\displaystyle\frac{\sum_{k\in\mathcal{K}_i^t}n_k\theta_{i,k}^{t+1}}{\sum_{k\in\mathcal{K}_i^t}n_k}, & \mathcal{K}_i^t\neq\varnothing,\\[7pt]
\theta_i^t, & \mathcal{K}_i^t=\varnothing.
\end{cases}
\end{equation}
Thus the denominator contains only clients that actually trained $\theta_i$. If every selected client covers the parameter, Eq.~(\ref{eq:coverage}) reduces to ordinary sample-weighted Federated Averaging (FedAvg); if no selected client covers it, the previous global value is retained. This prevents parameters in deeper prefixes from being diluted by clients that never trained those parameters. Conversely, parameters covered by fewer clients receive updates from a smaller effective participant set, so their optimization can be more sensitive to which eligible clients are selected in a round. 

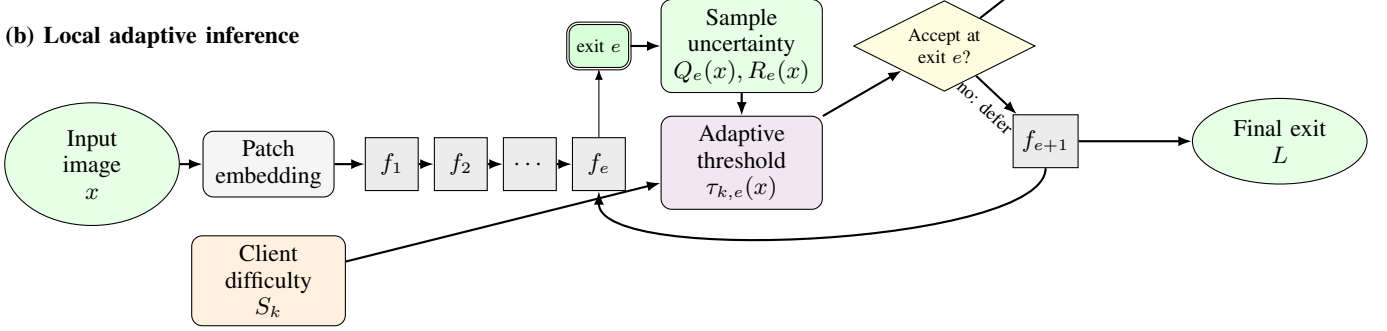
\begin{figure*}[!t]
\centering
\resizebox{\textwidth}{!}{%
\begin{tikzpicture}[node distance=3mm and 3mm,
box/.style={draw,rounded corners,align=center,minimum height=8mm,text width=17mm,font=\small},
server/.style={box,fill=blue!8},client/.style={box,fill=orange!12},
blk/.style={draw,minimum width=7mm,minimum height=7mm,font=\small,fill=gray!15},
exith/.style={draw,rounded corners,align=center,minimum width=7mm,minimum height=6mm,font=\scriptsize,fill=green!15,double,double distance=0.5pt},
term/.style={draw,ellipse,align=center,minimum height=7mm,text width=14mm,font=\small,fill=green!10},
arr/.style={-{Latex[length=1.8mm]},thick}]
\node[term] (start) {Global model\\$\boldsymbol{\theta}^{t}$};
\node[server,right=of start] (broadcast) {Server\\broadcast};
\node[client,right=4mm of broadcast,text width=16mm] (clientbox) {Selected\\client $k$};
\node[blk,right=6mm of clientbox] (b1) {$f_1$};
\node[blk,right=2mm of b1] (b2) {$f_2$};
\node[blk,right=2mm of b2] (b3) {$\cdots$};
\node[blk,right=2mm of b3] (bL) {$f_{m_k}$};
\node[exith,above=3mm of b1] (e1) {exit 1};
\node[exith,above=3mm of b2] (e2) {exit 2};
\node[exith,above=3mm of bL] (eL) {exit $m_k$};
\draw[arr] (b1)--(b2); \draw[arr] (b2)--(b3); \draw[arr] (b3)--(bL);
\draw[-{Latex[length=1.4mm]}] (b1)--(e1); \draw[-{Latex[length=1.4mm]}] (b2)--(e2); \draw[-{Latex[length=1.4mm]}] (bL)--(eL);
\node[font=\scriptsize,below=1mm of b2,xshift=4mm] {trainable subnet};
\node[server,right=6mm of bL,text width=16mm] (coverage) {Local update\\covered params};
\node[server,right=of coverage,text width=17mm] (agg) {Coverage-aware\\aggregation};
\node[term,right=of agg] (end) {Global model\\$\boldsymbol{\theta}^{t+1}$};
\draw[arr] (start)--(broadcast); \draw[arr] (broadcast)--(clientbox); \draw[arr] (clientbox)--(b1);
\draw[arr] (bL)--(coverage); \draw[arr] (coverage)--(agg); \draw[arr] (agg)--(end);
\draw[arr] (agg.south) -- ++(0,-8mm) -| node[below,font=\scriptsize]{next communication round} (broadcast.south);
\node[above=6mm of start,font=\bfseries\small] {(a) Federated training};
\end{tikzpicture}}
\vspace{1mm}
\resizebox{\textwidth}{!}{%
\begin{tikzpicture}[node distance=3mm and 3mm,
box/.style={draw,rounded corners,align=center,minimum height=8mm,text width=17mm,font=\small},
client/.style={box,fill=orange!12},sample/.style={box,fill=green!10},
blk/.style={draw,minimum width=7mm,minimum height=7mm,font=\small,fill=gray!15},
exith/.style={draw,rounded corners,align=center,minimum width=7mm,minimum height=6mm,font=\scriptsize,fill=green!15,double,double distance=0.5pt},
term/.style={draw,ellipse,align=center,minimum height=7mm,text width=14mm,font=\small,fill=green!10},
dec/.style={draw,diamond,align=center,inner sep=0.5pt,text width=14mm,font=\scriptsize,fill=yellow!15,aspect=2},
arr/.style={-{Latex[length=1.8mm]},thick}]
\node[term] (input) {Input image\\$x$};
\node[box,right=of input,fill=gray!8,text width=15mm] (patch) {Patch\\embedding};
\node[blk,right=4mm of patch] (b1) {$f_1$}; \node[blk,right=2mm of b1] (b2) {$f_2$};
\node[blk,right=2mm of b2] (b3) {$\cdots$}; \node[blk,right=2mm of b3] (be) {$f_e$};
\node[exith,above=9mm of be] (eh) {exit $e$};
\node[client,below=5mm of patch,text width=18mm] (cprofile) {Client difficulty\\$S_k$};
\node[sample,right=4mm of eh,text width=19mm] (unc) {Sample uncertainty\\$Q_e(x),R_e(x)$};
\node[box,below=3mm of unc,fill=violet!10,text width=19mm] (thresh) {Adaptive threshold\\$\tau_{k,e}(x)$};
\node[dec,right=4mm of unc] (decide) {Accept at\\exit $e$?};
\node[term,above right=6mm and 3mm of decide] (predict) {Return\\$\hat y_e(x)$};
\node[blk,below right=6mm and 3mm of decide] (nextexit) {$f_{e+1}$};
\node[term,right=15mm of nextexit] (fallback) {Final exit\\$L$};
\draw[arr] (input)--(patch); \draw[arr] (patch)--(b1); \draw[arr] (b1)--(b2); \draw[arr] (b2)--(b3); \draw[arr] (b3)--(be);
\draw[-{Latex[length=1.4mm]}] (be)--(eh); \draw[arr] (cprofile)--(thresh); \draw[arr] (eh)--(unc);
\draw[arr] (unc)--(thresh); \draw[arr] (thresh)--(decide);
\draw[arr] (decide)--node[above,font=\scriptsize,sloped]{yes}(predict);
\draw[arr] (decide)--node[below,font=\scriptsize,sloped]{no: defer}(nextexit);
\draw[arr] (nextexit.south)..controls +(0,-10mm) and +(0,-10mm)..(be.south); \draw[arr] (nextexit)--(fallback);
\node[above=6mm of input,xshift=8mm,font=\bfseries\small] {(b) Local adaptive inference};
\end{tikzpicture}}
\caption{Federated training (a) and local client inference (b).}
\label{fig:framework}
\end{figure*}

\subsection{Communication and Computation Metrics}
Let $B_{k,t}$ be the number of model bytes transmitted in one direction for client $k$ during round $t$. Because the experiments count both server-to-client broadcast and client-to-server upload, cumulative model communication is
\begin{equation}
 \mathrm{Comm}=2\sum_{t=0}^{T_{\mathrm{FL}}-1}\sum_{k\in\mathcal{S}_t}B_{k,t},
\end{equation}
excluding protocol overhead. With full-model participation, $B_{k,t}$ is the serialized full-model size; with a subnet it is the serialized active prefix size.

For inference, the paper reports a depth-based computation proxy rather than hardware-specific latency. Let $N$ be the number of evaluation samples and let $e_i$ be the exit index used by sample $i$. Average exit depth and the percentage depth saving relative to evaluating all $L$ stages are
\begin{equation}
 \bar e=\frac{1}{N}\sum_{i=1}^{N}e_i,\qquad
 \mathrm{Saved}=100\left(1-\frac{\bar e}{L}\right).
\label{eq:saving}
\end{equation}
This metric measures skipped Transformer stages; it should not be interpreted as an exact floating-point-operation (FLOP) or wall-clock reduction.

\section{Experimental Results and Analysis}
\label{sec:results}
\subsection{Datasets and Experimental Setup}
We evaluate CIFAR-10 and CIFAR-100, which contain 10 and 100 image classes respectively \cite{krizhevsky2009}, and ImageNet-1K, which contains 1000 classes \cite{deng2009imagenet}. An \emph{IID} partition means client data are independently and identically distributed from the same class distribution. For non-IID experiments, class-wise client proportions are sampled from a symmetric Dirichlet distribution with concentration parameter $\alpha$; smaller $\alpha$ produces stronger label skew because each class is concentrated on fewer clients.

All experiments use the same six-exit ViT family. CIFAR-10 uses five local epochs, the AdamW optimizer with cosine learning-rate scheduling, standard MixUp/CutMix and RandAugment augmentation, self-distillation, and a 200-round budget. It includes IID experiments with 3, 6, and 9 clients, plus a six-client Dirichlet sweep with $\alpha\in\{0.5,0.3,0.1,0.05\}$. CIFAR-100 uses six exits with embedding dimension 256 and four attention heads, five local epochs, batch size 64, AdamW with cosine scheduling, MixUp, RandAugment, stochastic-depth regularization of 0.1, and self-distillation weight 0.1 with $T_{\mathrm{KD}}=3$. Its client-count sweep uses $2,4,6,8,10$ clients at $\alpha=0.1$, and its heterogeneity sweep uses six clients with the same four $\alpha$ values.

Adversarial routing is tested with Fast Gradient Sign Method (FGSM) and 10-step Projected Gradient Descent (PGD) \cite{goodfellow2015,madry2018}. The symbol $\epsilon$ denotes the $\ell_\infty$ perturbation budget, i.e., the maximum absolute element-wise input change, with $\epsilon\in\{10^{-4},10^{-3},10^{-2}\}$. These attacks are generated against the deepest exit and are used for within-run policy comparisons; no adversarial training is applied. 

\subsection{Architecture Instantiation and Metrics}
The implementation starts from the \texttt{timm} checkpoint \path{vit_small_patch16_224.augreg_in21k_ft_in1k}, which is a small ViT with $16\times16$ patches (ViT-S/16) trained with augmentation and regularization (AugReg), pretrained on ImageNet-21k and fine-tuned on ImageNet-1k. We retain the first six Transformer blocks. CIFAR images are resized to $224\times224$ using bicubic interpolation, matching the checkpoint configuration. At exit $e$, let $z_e^{\mathrm{cls}}$ be the class token, $z_{e,j}$ the $j$th patch token, and $N_p$ the number of patch tokens. The exit representation averages the class token and mean patch token, applies layer normalization $\operatorname{LN}(\cdot)$, and reuses exit-head parameters $(W_e,b_e)$ from Section~III-A:
\begin{equation}
 p_e=\operatorname{softmax}\!\left(W_e\operatorname{LN}\!\left[\tfrac12\left(z_e^{\mathrm{cls}}+\tfrac1{N_p}\sum_{j=1}^{N_p}z_{e,j}\right)\right]+b_e\right).
\end{equation}
Top-1 accuracy is the fraction of samples whose most probable class is correct; top-5 accuracy counts a sample as correct when its label is among the five most probable classes. The computation metric is Eq.~(\ref{eq:saving}); communication counts both transfer directions as defined above.

\subsection{CIFAR-10 and ImageNet-1K Coverage}
With 3, 6, and 9 IID CIFAR-10 clients, global test accuracy is 90.62\%, 90.17\%, and 89.94\%, while mean client-local accuracy is 91.78\%, 91.30\%, and 91.02\%, respectively. Under six-client Dirichlet partitions, global test accuracy decreases from 88.90\% at $\alpha=0.5$ to 79.50\% at $\alpha=0.05$, consistent with stronger label skew.

\begin{table}[!t]
\caption{CIFAR-10 Evaluation Across Client Counts and Exit Policies}
\label{tab:cifar10}
\centering\scriptsize
\begin{tabular}{lccc}\toprule
Setting & Global acc. (\%) & Mean client (\%) & Saved (\%)\\\midrule
3 clients, IID & 90.62 & 91.78 & --\\
6 clients, IID & 90.17 & 91.30 & --\\
9 clients, IID & 89.94 & 91.02 & --\\\midrule
Full depth & 90.17 & -- & 0.0\\
Fixed threshold & 87.00 & -- & 37.5\\
Global resource-aware & 87.60 & -- & 36.3\\
Robustness-aware & 88.70 & -- & 32.0\\\bottomrule
\end{tabular}
\end{table}
Table~\ref{tab:cifar10} isolates the CIFAR-10 accuracy--depth trade-off. Robustness-aware routing reaches 88.70\% global test accuracy with 32.0\% depth saving, whereas the fixed threshold reaches 87.00\% with 37.5\% saving. The additional uncertainty signals therefore recover 1.70 percentage points (pp) of accuracy while using 5.5 pp more of the six-stage depth budget.

ImageNet-1K tests whether the same routing design transfers beyond CIFAR. The six-client full-depth model reaches 75.4\% top-1 and 92.3\% top-5 accuracy. Fixed and global resource-aware routing reach 72.1\% and 72.6\% top-1 accuracy with 34\% and 33\% depth saving, respectively. Robustness-aware routing reaches 73.7\% top-1 with 30\% saving (Table~\ref{tab:imagenet}).

\begin{table}[!t]
\caption{Six-Client ImageNet-1K Accuracy--Depth Trade-off}
\label{tab:imagenet}
\centering\footnotesize
\begin{tabular}{lccc}\toprule
Policy & Top-1 (\%) & Avg. depth & Saved (\%)\\\midrule
Full depth & 75.4 & 6.00 & 0\\
Fixed threshold & 72.1 & 3.96 & 34\\
Global resource-aware & 72.6 & 4.02 & 33\\
Robustness-aware & 73.7 & 4.20 & 30\\\bottomrule
\end{tabular}
\end{table}

\subsection{Matched Comparison with Established Methods}
Table~\ref{tab:prior} compares CIFAR-100 accuracy using six clients, Dirichlet $\alpha=0.1$, the same training budget, and the same evaluation split. The reproduced HeteroFL, DepthFL, ScaleFL, and ReeFL runs reach 68.20\%, 70.10\%, 71.30\%, and 71.80\%, respectively; our full-depth model reaches 72.28\%. Thus the full-depth accuracy advantage over the strongest reproduced baseline, ReeFL, is 0.48 pp. The comparison is matched with respect to client count, Dirichlet partition setting, training budget, and evaluation split. The available experiment record does not establish that all baseline runs used identical backbone initialization/pretraining, augmentation, self-distillation, or method-specific tuning. 

A common matched FLOP/latency trace is unavailable for all reproduced baselines, so Table~\ref{tab:prior} reports each method's compute-control mechanism rather than inventing a cross-method saving percentage. Table~\ref{tab:tradeoff} contains our measured depth proxy. For context, ScaleFL reports up to $2\times$ latency and $4\times$ model-size reduction under its own protocol \cite{ilhan2023scalefl}; those values are not mixed with our matched runs.

\begin{table}[!t]
\caption{Matched CIFAR-100 Comparison: 6 Clients, Dirichlet $\alpha=0.1$}
\label{tab:prior}
\centering\scriptsize
\setlength{\tabcolsep}{2.6pt}
\begin{tabular}{lccp{2.4cm}}\toprule
Method & Acc. (\%) & $\Delta$ vs. ours (pp) & Compute control\\\midrule
HeteroFL \cite{diao2021} & 68.20 & -4.08 & model width\\
DepthFL \cite{kim2023} & 70.10 & -2.18 & trainable depth\\
ScaleFL \cite{ilhan2023scalefl} & 71.30 & -0.98 & width + depth\\
ReeFL \cite{lee2024} & 71.80 & -0.48 & recurrent exits\\
Ours, full depth & 72.28 & 0.00 & 6 stages\\
Ours, robustness-aware & 70.01 & -2.27 & avg. depth 4.05\\\bottomrule
\end{tabular}
\end{table}

\begin{figure*}[!t]
\centering
\includegraphics[width=0.94\textwidth]{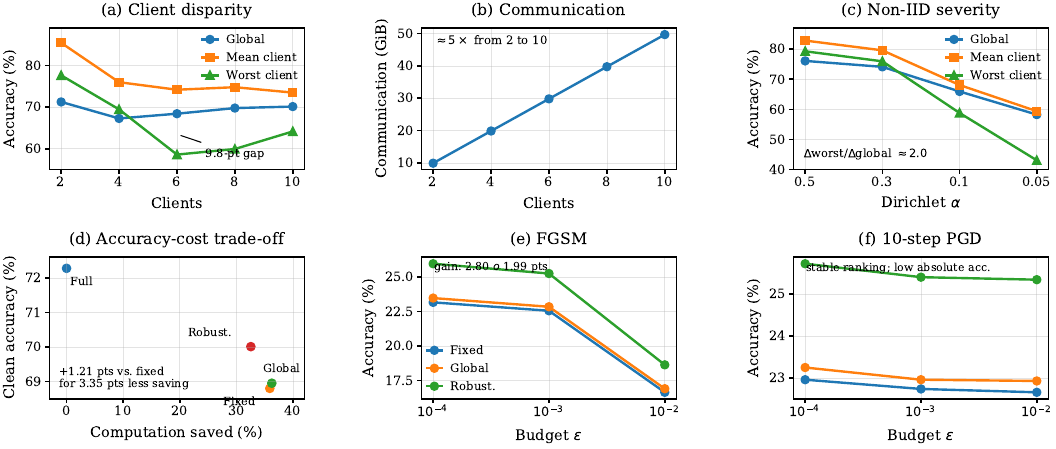}
\caption{CIFAR-100 analysis: client disparity and communication (a--b), non-IID degradation and clean accuracy--depth trade-off (c--d), and FGSM/PGD routing comparisons (e--f).}
\label{fig:results}
\end{figure*}

\subsection{CIFAR-100 Scalability, Heterogeneity, and Efficiency}
Figure~\ref{fig:results}(a) shows that global test accuracy changes by 1.12 pp between two and ten clients, whereas worst-client accuracy reaches 58.61\% at six clients, 9.83 pp below the corresponding global test accuracy. Cumulative two-way model communication grows from 9.93 to 49.66~GiB (gibibytes) as the client count increases from two to ten [Fig.~\ref{fig:results}(b)], consistent with the communication definition in Section~III-G. Under six-client Dirichlet partitions, reducing $\alpha$ from 0.50 to 0.05 lowers global accuracy by 17.85 pp and worst-client accuracy by 36.23 pp [Fig.~\ref{fig:results}(c)]. The stronger degradation of the worst client motivates replacing one global threshold with client-dependent acceptance thresholds.

Table~\ref{tab:tradeoff} and Fig.~\ref{fig:results}(d) compare routing policies on the same CIFAR-100 global test split. Fixed and global resource-aware routing save about 36\% of stage depth but lose about 3.4 pp relative to full-depth inference. Robustness-aware routing reaches 70.01\% at average depth 4.05, corresponding to 32.55\% depth saving. Relative to the fixed threshold, it recovers 1.21 pp of accuracy while using 3.35 pp more of the depth budget.

\begin{table}[!t]
\caption{CIFAR-100 Global-Test Accuracy--Depth Trade-off}
\label{tab:tradeoff}
\centering\footnotesize
\begin{tabular}{lccc}\toprule
Policy & Acc. (\%) & Avg. depth & Saved (\%)\\\midrule
Full depth & 72.28 & 6.00 & 0.00\\
Fixed threshold & 68.80 & 3.85 & 35.90\\
Global resource-aware & 68.95 & 3.83 & 36.25\\
Robustness-aware & 70.01 & 4.05 & 32.55\\\bottomrule
\end{tabular}
\end{table}

\subsection{Routing-Term and Adversarial Analysis}
The fixed policy uses none of the adaptive terms in Eq.~(\ref{eq:threshold-family}). The global resource-aware policy activates $Q_e(x)$ and depth relaxation, changing clean accuracy by only +0.15 pp and depth saving by +0.35 pp relative to fixed routing. The robustness-aware policy additionally activates $0.20R_e(x)$; clean accuracy increases by a further 1.06 pp while depth saving decreases by 3.70 pp. In other words, the extra uncertainty term deliberately defers more ambiguous samples and trades some computation saving for higher accuracy.

Under FGSM, robustness-aware routing exceeds fixed-threshold accuracy by 2.80, 2.70, and 1.99 pp across the three $\epsilon$ values [Fig.~\ref{fig:results}(e)]. Under 10-step PGD, the corresponding gains are 2.76, 2.66, and 2.68 pp [Fig.~\ref{fig:results}(f)]. These are within-run routing comparisons, not evidence of certified robustness; the absolute attack accuracy also depends on the attack implementation and model normalization.

\section{Conclusion}
This paper studied local early-exit inference for heterogeneous FL and asked whether decisions should depend on both the client and the current sample. The framework combines a client-specific base threshold from label entropy and shallow-exit confidence with sample confidence, predictive entropy, class margin, and relative exit depth; coverage-aware aggregation averages a parameter only across clients that trained it. The conditional-error analysis shows that deferral helps only when the deeper exit is more accurate on the deferred subset. Across CIFAR-10, CIFAR-100, and ImageNet-1K, this produces a consistent accuracy--depth trade-off: robustness-aware routing reaches 88.70\% on CIFAR-10 with 32.0\% depth saving; CIFAR-100 full-depth accuracy is 72.28\%, 0.48 pp above the strongest matched baseline (ReeFL at 71.80\%), while robustness-aware routing reaches 70.01\% with 32.55\% saving; and ImageNet-1K reaches 73.7\% top-1 with 30\% saving versus 75.4\% at full depth. Under Fast Gradient Sign Method (FGSM) and 10-step Projected Gradient Descent (PGD), robustness-aware routing also stays above fixed-threshold routing in the tested settings. The remaining limitations are matched hardware latency/FLOP evaluation and broader adversarial suites. 

\section*{ACKNOWLEDGMENT}
\small
This work was supported in part by the Estonian Research Council grant PUT PRG1467 ``CRASHLESS'', and EU Grant Project 101160182 ``TAICHIP''.

\bibliographystyle{IEEEtran}
\bibliography{references}
\end{document}